\documentclass[11pt,a4paper]{article}

\usepackage[utf8]{inputenc}
\usepackage[T1]{fontenc}
\usepackage{mathptmx}           
\usepackage[margin=1in]{geometry}
\usepackage{hyperref}
\usepackage{url}
\usepackage{booktabs}
\usepackage{amsfonts}
\usepackage{amsmath}
\usepackage{amssymb}
\usepackage{graphicx}
\usepackage{xcolor}
\usepackage{multirow}
\usepackage{enumitem}

\hypersetup{
    colorlinks=true,
    linkcolor=blue,
    citecolor=blue,
    urlcolor=blue,
}

\title{\vspace{-1.5cm}\textbf{Critical Damping as a Momentum Schedule:\\Multi-Seed Validation, a Hybrid Recipe, and an Exhaustive\\Negative Result on Surgical Layer Selection}}

\author{Ivan Pasichnyk\\
We Label Data Inc.\\
\texttt{ivan@welabeldata.com}\\
\url{https://orcid.org/0009-0004-8154-3345}}

\date{}

\begin{document}

\maketitle

\begin{abstract}
The critical damping condition of the damped harmonic oscillator model of SGD with momentum (Qian, 1999) yields a momentum schedule with no tuned hyperparameters: $\mu(t) = 1 - 2\sqrt{\alpha(t)}$ (a fixed, never-tuned stability clamp aside). We validate this schedule across five random seeds on ResNet-18/CIFAR-10 (200-epoch cosine learning-rate schedule): it reaches 90\% test accuracy $2.34\times$ faster than constant $\mu=0.9$ (range $1.71$--$2.86\times$, 5/5 seeds, one-sided paired $t$-test $p=4\times10^{-4}$), at the cost of a real final-accuracy deficit of $0.46$~pp ($5/5$ seeds, $p=0.009$). A short-schedule control shows the speedup is not a schedule-length artifact: baselines with the cosine schedule compressed to 60/100 epochs either pay $0.5$--$0.9$~pp of final accuracy or remain slower to 90\% than the physics schedule at equal accuracy---physics Pareto-dominates the accuracy-matched compressed baseline. A hybrid recipe---critical-damping momentum until 90\% accuracy, then constant $\mu=0.9$---removes the deficit while keeping the speedup: $95.45 \pm 0.05\%$ final accuracy (indistinguishable from baseline) at $2.4\times$ faster progress to 90\% ($n=5$). The speedup generalizes across architectures---VGG-16 without skip connections reaches 90\% test accuracy $1.72\times$ faster ($n=3$)---while on CIFAR-100 the effect is smaller: early-training gains persist ($2$--$4\times$ to mid-training thresholds) but the pure schedule pays a larger final-accuracy cost ($-1.7$~pp), compressing the accuracy-matched gain to $1.14\times$ and delimiting the schedule's scope.

We also report an exhaustive negative result on surgical layer selection. Version~2 of this paper claimed that gradient attribution computed on misclassified images selects \emph{which} layers to retrain. Running the identical surgical-correction protocol on \textbf{all 35 combinations} of 3-of-7 layer groups shows the diagnostically selected triple ranks 11th of 35 (exact $p=0.31$)---no better than random selection. What survives is weaker but real: combinations containing the top-ranked layer outperform those without it ($+6.2$ vs $-1.4$ mean net error reduction), and the \emph{bottom} of the gradient-norm ranking reliably predicts the most harmful interventions (down to $-20$ net errors). Gradient attribution on errors is a \emph{harm-avoidance} signal, not an optimizer of layer selection. We release the full 35-combination landscape as a baseline we believe should accompany any layer-selection claim.
\end{abstract}


\section{Introduction}

Momentum and learning rate are usually tuned as independent hyperparameters. The damped harmonic oscillator model of SGD with momentum~\cite{qian1999momentum} says they should not be: for any learning-rate schedule $\alpha(t)$, the critical damping condition fixes the momentum trajectory completely, $\mu(t) = 1 - 2\sqrt{\alpha(t)}$, with no tuned hyperparameters (the implementation applies a fixed, never-tuned stability clamp; Section~\ref{sec:multiseed}). This paper asks two questions about that formula, and answers a third one we did not plan to ask.

\textbf{Does it work?} Yes, with a quantified trade-off (Section~\ref{sec:multiseed}). Across five random seeds on ResNet-18/CIFAR-10 under a fixed 200-epoch cosine schedule, critical-damping momentum reaches 90\% test accuracy $2.34\times$ faster than the standard constant $\mu = 0.9$ (range $1.71$--$2.86\times$, 5/5 seeds, one-sided paired $t$-test $p=4\times10^{-4}$). The price is real, not seed noise: a final-accuracy deficit of $0.46$~pp ($5/5$ seeds, $p = 0.009$). A short-schedule control (Section~\ref{sec:control}) rules out the schedule-length explanation: physics is faster to 90\% than any compressed baseline of equal final accuracy.

\textbf{Can the trade-off be removed?} Yes (Section~\ref{sec:hybrid}). A hybrid recipe---critical damping until 90\% accuracy, then constant $\mu=0.9$---keeps the early speedup and recovers full final accuracy: across five seeds, Hybrid-90 reaches $95.45 \pm 0.05\%$ (statistically indistinguishable from the baseline's $95.49 \pm 0.08\%$) while hitting 90\% in $45.2 \pm 9.5$ epochs---$2.4\times$ faster than baseline, matching pure critical damping. The oscillator intuition explains why: near-critical damping is optimal for fast descent in the early high-curvature landscape, while the underdamped regime's oscillation aids late-stage exploration.

\textbf{Does gradient attribution on errors select which layers to repair?} No---and version~2 of this paper claimed it does (Section~\ref{sec:exhaustive}). We ran the identical surgical-correction protocol on \emph{all 35 combinations} of 3-of-7 ResNet-18 layer groups. The diagnostically selected triple ranks 11th of 35 (exact $p = 0.31$). Two weaker signals survive: combinations containing the top-ranked layer outperform those without it ($+6.2$ vs $-1.4$ mean net error reduction), and the bottom of the gradient-norm ranking reliably identifies the most \emph{harmful} interventions ($-13$ to $-20$ net errors). Gradient attribution on errors is a harm-avoidance signal, not a selector of optimal repair targets.

Contributions, in order of the strength of the evidence:
\begin{enumerate}[leftmargin=*,topsep=2pt,itemsep=1pt]
    \item \textbf{Multi-seed empirical validation} of the tuning-free critical-damping schedule: $2.34\times$ faster to 90\% ($n=5$, one-sided $p=4\times10^{-4}$), with an honest quantification of its final-accuracy cost ($-0.46$~pp, $p=0.009$). The formula is Qian's~\cite{qian1999momentum}; the controlled validation and the trade-off quantification are ours.
    \item \textbf{The Hybrid-90 recipe}: a practical schedule that dominates both of its parents (fast early convergence of critical damping, full final accuracy of constant momentum).
    \item \textbf{An exhaustive negative result} on error-gradient layer selection, with the full 35-combination landscape released as a baseline. To our knowledge no prior surgical fine-tuning work reports its selection rule against the complete combinatorial null.
\end{enumerate}

\section{Related Work}

\paragraph{ODE models of momentum.} Qian~\cite{qian1999momentum} showed that gradient descent with momentum, in the continuous-time limit near a local minimum, is equivalent to a set of coupled damped harmonic oscillators, and that optimal convergence corresponds to critical damping of each eigencomponent. Su et al.~\cite{su2016differential} derived an ODE for Nesterov's accelerated gradient method. Shi et al.~\cite{shi2022understanding} introduced high-resolution ODEs distinguishing Nesterov's method from heavy-ball momentum. \emph{Our work builds on Qian's model: we derive the critical damping condition as a practical schedule and, more importantly, use damping-regime analysis as the foundation for a diagnostic pipeline.}

\paragraph{Learning rate--momentum coupling.} Smith~\cite{smith2017cyclical,smith2018superconvergence} empirically discovered that coupling momentum inversely to learning rate improves convergence (the 1cycle policy). \emph{Difference from our work:} 1cycle approximates the critical damping curve with a piecewise-linear schedule but does not derive the relationship from physics and does not use the framework for diagnostics.

\paragraph{Physics-informed optimization.} Karoni et al.~\cite{karoni2026adaptive} independently derive a momentum--learning-rate coupling from Hamiltonian mechanics, obtaining per-parameter adaptive damping (iKFAD, Cubic Damping, CADAM). Their framework is more general: per-parameter friction, cubic damping terms, bridges the Adam--mSGD gap on Transformers. \emph{Difference:} their work is an optimizer; ours is a diagnostic. We compare directly in Section~\ref{sec:exp3} and find that layer-level correction (ours and iKFAD-guided) outperforms their per-parameter correction.

\paragraph{Adaptive methods.} Adam~\cite{kingma2015adam} and AdamW~\cite{loshchilov2019decoupled} adapt the effective learning rate per parameter using gradient statistics. Polyak~\cite{polyak1964some} introduced the heavy-ball method; Nesterov~\cite{nesterov1983method} proposed a lookahead variant; Sutskever et al.~\cite{sutskever2013importance} established $\mu=0.9$ as a practical default. Wang et al.~\cite{cutkosky2024marginal} show that momentum's marginal value diminishes for small learning rates---consistent with our analysis where $\mu_c \to 1$ as $\alpha \to 0$. Closest to our schedule are momentum \emph{tuners and schedules}: YellowFin~\cite{zhang2017yellowfin} tunes momentum and learning rate jointly from a local quadratic model with a closed-loop controller; Sun et al.~\cite{sun2021adaptive} derive an adaptive momentum from quadratic optimization; Demon~\cite{chen2019demon} decays momentum over training by a fixed rule. \emph{Difference:} all three add machinery (estimators, controllers, or decay hyperparameters), whereas Eq.~\ref{eq:main} is a closed-form function of the existing learning-rate schedule with nothing to estimate or tune. We also note the classical exact result for quadratics---the optimal discrete momentum $\mu^* = (1-\sqrt{\alpha\lambda})^2 = 1 - 2\sqrt{\alpha\lambda} + \alpha\lambda$---of which our schedule is the first-order (small-$\alpha$) approximation at unit curvature; and concurrent work on critically damped momentum near interpolation~\cite{muscarnera2026critical}.

\paragraph{Knowledge editing and model repair.} Meng et al.~\cite{meng2022locating,meng2023memit} developed ROME and MEMIT for locating and editing factual associations in GPT models by modifying MLP weights in specific layers. \emph{Difference:} they edit specific facts via causal tracing; we diagnose general error patterns via gradient norms on misclassified inputs. Zou et al.~\cite{zou2023representation} proposed Representation Engineering, which reads and controls high-level concepts via population-level representations. \emph{Difference:} they identify \emph{directions} in activation space; we identify \emph{layers} where errors concentrate. Lee et al.~\cite{lee2023surgical} showed that surgical fine-tuning of a subset of layers can match or outperform full fine-tuning for distribution shifts. \emph{Difference:} they showed \emph{that} surgical fine-tuning works; we test a principled selection rule for \emph{which layers to target} and show, via an exhaustive 35-combination baseline (Section~\ref{sec:exhaustive}), that it does not outperform random selection --- the rule's reliable signal is harm avoidance. Recent work on hierarchical alignment~\cite{hierarchical2025} decomposes objectives across functionally specialized layer blocks in LLMs.

\section{Background: The Oscillator Model}

This section reviews known results~\cite{qian1999momentum,su2016differential,shi2022understanding} and derives the explicit schedule used throughout this paper. The oscillator model itself is not our contribution; we include the derivation to set notation.

\subsection{SGD with Momentum as a Damped Oscillator}

SGD with momentum updates parameters $\theta$ via:
\begin{align}
v_{t+1} &= \mu\, v_t - \alpha\, \nabla \mathcal{L}(\theta_t), \label{eq:sgd_v}\\
\theta_{t+1} &= \theta_t + v_{t+1}, \label{eq:sgd_theta}
\end{align}
where $v_t$ is the velocity buffer, $\alpha$ is the learning rate, $\mu$ is the momentum coefficient, and $\nabla\mathcal{L}$ is the loss gradient.

Eliminating $v_t$ by substituting Eq.~\ref{eq:sgd_v} into Eq.~\ref{eq:sgd_theta} and taking the continuous-time limit, we obtain the equation of a forced damped harmonic oscillator~\cite{qian1999momentum}:
\begin{equation}
\ddot{x} + \gamma\,\dot{x} + \omega^2\, x = F(t),
\label{eq:oscillator}
\end{equation}
with the correspondence shown in Table~\ref{tab:correspondence}.

\begin{table}[ht]
\centering
\caption{Correspondence between the damped harmonic oscillator and SGD with momentum~\cite{qian1999momentum}.}
\label{tab:correspondence}
\begin{tabular}{@{}ll@{}}
\toprule
\textbf{Oscillator} & \textbf{SGD with Momentum} \\
\midrule
Position $x$ & Parameters $\theta$ \\
Velocity $\dot{x}$ & Velocity buffer $v$ \\
Damping coefficient $\gamma$ & $1 - \mu$ \\
Natural frequency $\omega$ & $\propto \sqrt{\alpha}$ \\
External force $F(t)$ & stochastic/anisotropic gradient residual (see text) \\
\bottomrule
\end{tabular}
\end{table}

The second-order recurrence obtained by eliminating $v_t$ from Eqs.~\ref{eq:sgd_v}--\ref{eq:sgd_theta} is:
\begin{equation}
\theta_{t+1} - (1+\mu)\,\theta_t + \mu\,\theta_{t-1} = -\alpha\,\nabla\mathcal{L}(\theta_t).
\label{eq:second_order}
\end{equation}
Near a minimum $\theta^*$ we linearize $\nabla\mathcal{L}(\theta) \approx H(\theta - \theta^*) + r(t)$, where $H$ is the local Hessian and $r(t)$ collects the stochastic and anisotropic residual. For an eigenmode of $H$ with curvature $\lambda$, dividing by $\Delta t^2$ and identifying $\gamma = (1-\mu)/\Delta t$ and $\omega^2 = \lambda\alpha/\Delta t^2$ recovers Eq.~\ref{eq:oscillator}, with the residual $r(t)$ playing the role of the external force $F(t)$. Setting $\Delta t = 1$ and unit curvature $\lambda = 1$ gives $\gamma = 1 - \mu$ and $\omega = \sqrt{\alpha}$; Section~\ref{sec:spectral-grounding} discusses the general $\lambda_k$ case. The continuous-time identification is a consistent limit when $1-\mu = O(\Delta t)$ and $\alpha = O(\Delta t^2)$, i.e.\ in the heavy-momentum, small-step regime typical of the schedules studied here.

\begin{figure}[t]
\centering
\includegraphics[width=\textwidth]{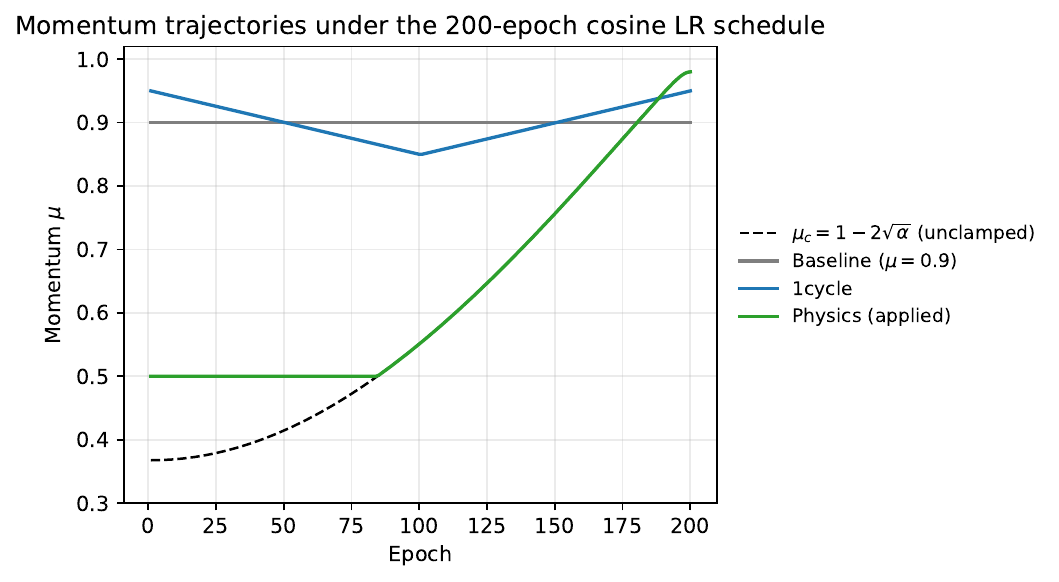}
\caption{Momentum trajectories for the three methods under the 200-epoch cosine learning-rate schedule. The black dashed line is the unclamped critical-damping curve $\mu_c = 1 - 2\sqrt{\alpha}$. The physics method (green, applied momentum) is clamped at $0.5$ through epoch~84 and tracks the critical curve thereafter.}
\label{fig:momentum}
\end{figure}

\subsection{Three Damping Regimes}

The behavior of a damped oscillator depends on the discriminant $\Delta = \gamma^2 - 4\omega^2$:
\begin{itemize}[leftmargin=*,topsep=2pt,itemsep=1pt]
    \item \textbf{Underdamped} ($\gamma < 2\omega$, $\Delta < 0$): the system oscillates around equilibrium before converging. In optimization, this manifests as fluctuating loss and accuracy curves.
    \item \textbf{Overdamped} ($\gamma > 2\omega$, $\Delta > 0$): the system approaches equilibrium without oscillation but unnecessarily slowly.
    \item \textbf{Critically damped} ($\gamma = 2\omega$, $\Delta = 0$): the unique regime that reaches equilibrium in minimum time without oscillation.
\end{itemize}

\subsection{The Critical Damping Schedule}

Setting $\gamma = 2\omega$ and substituting $\gamma = 1 - \mu$ and $\omega = \sqrt{\alpha}$:
\begin{equation}
1 - \mu = 2\sqrt{\alpha} \quad\Longrightarrow\quad \boxed{\mu(t) = 1 - 2\sqrt{\alpha(t)}}.
\label{eq:main}
\end{equation}

Given \emph{any} learning rate schedule $\alpha(t)$, Eq.~\ref{eq:main} determines a momentum trajectory that targets critical damping throughout training. We call this \emph{$\beta$-scheduling}.

\subsection{Spectral Grounding}
\label{sec:spectral-grounding}

Equation~\ref{eq:main} sets momentum for a single-mode oscillator with unit curvature. A neural network loss has a spectrum of local curvatures~$\{\lambda_k\}$, so one might ask why a global schedule works at all. Recent work by Olsen et al.~\cite{olsen2026spectra} derives the stochastic dynamics of the squared singular values of weight matrices under SGD, showing that their stationary distribution is heavy-tailed with mass concentrated around a dominant scale that evolves slowly relative to the training trajectory. This gives a principled answer: our global $\mu(t) = 1 - 2\sqrt{\alpha(t)}$ critically damps the \emph{dominant} spectral mode, while sub-dominant modes are left mildly under- or overdamped. The empirical robustness of Eq.~\ref{eq:main} across runs and architectures reflects this separation of scales rather than a coincidence.

The natural refinement is a \emph{per-eigenmode} schedule $\mu_k(t) = 1 - 2\sqrt{\lambda_k(t) \cdot \alpha(t)}$, which critically damps every mode simultaneously. Implementing this requires an online estimate of the Hessian spectrum --- tractable via Lanczos or Hessian-vector products, but nontrivial at scale. We defer the per-eigenmode construction and its joint analysis with Olsen et al.'s SDE to future work (see also Karoni et al.~\cite{karoni2026adaptive} for a related per-parameter perspective). For the remainder of this paper, the global schedule is sufficient: the results in Sections~\ref{sec:exp1} onward show that damping the dominant mode already captures the phenomena of interest, without tuning $\mu$ per layer or per parameter.

\subsection{Qualitative Predictions}

The critical damping condition makes three testable predictions:
\begin{enumerate}[leftmargin=*,topsep=2pt,itemsep=1pt]
    \item \textbf{Fastest early convergence.} Critically damped systems have the shortest settling time.
    \item \textbf{No oscillation.} Training curves should improve monotonically during the high-LR phase.
    \item \textbf{Competitive final accuracy.} The oscillator model captures macroscopic dynamics but not the fine structure of the loss landscape. Final accuracy should be comparable but not necessarily superior.
\end{enumerate}

\subsection{Retroactive Explanations of Empirical Heuristics}

Equation~\ref{eq:main} retroactively explains several established practices:
\begin{itemize}[leftmargin=*,topsep=2pt,itemsep=1pt]
    \item \textbf{Why $\mu = 0.9$ is a common default.} At $\alpha = 0.01$: $\mu_c = 1 - 2\sqrt{0.01} = 0.8$. The value 0.9 is close---slightly underdamped but workable.
    \item \textbf{Why 1cycle couples momentum inversely to LR.} Higher $\alpha$ requires lower $\mu$ for critical damping. The 1cycle policy~\cite{smith2018superconvergence} approximates this curve.
    \item \textbf{Why learning rate warmup helps.} Starting with small $\alpha$ gives $\mu_c \approx 1$ (near-zero damping), allowing velocity to build smoothly.
\end{itemize}

\section{Experiment 1: Multi-Seed Validation of the Schedule}
\label{sec:multiseed}
\label{sec:exp1}

We test the oscillator model's qualitative predictions across five random seeds (42, 123, 456, 789, 1337). Runs are not bit-deterministic (cuDNN autotuning); repeated runs of the same configuration and seed can differ by a few hundredths of a pp in accuracy and a few epochs in milestones, which accounts for small discrepancies between experiments that reuse a configuration. The headline result is convergence speed ($2.34\times$ to 90\%, all seeds); the schedule also carries a real final-accuracy cost that single-seed experiments cannot distinguish from noise---we quantify both.

\subsection{Multi-Seed Results}

\begin{table}[ht]
\centering
\caption{Multi-seed results ($n=5$, mean $\pm$ std): ResNet-18/CIFAR-10, 200 epochs, cosine LR $0.1 \to 10^{-4}$. Nesterov and heavy-ball results (3 seeds) in the text.}
\label{tab:multiseed}
\begin{tabular}{@{}lcc@{}}
\toprule
\textbf{Method} & \textbf{Best Test Accuracy} & \textbf{Epochs to 90\%} \\
\midrule
Baseline ($\mu=0.9$) & \textbf{95.49 $\pm$ 0.08\%} & 107.8 $\pm$ 10.1 \\
1cycle (Smith) & 95.40 $\pm$ 0.05\% & 101.4 $\pm$ 16.0 \\
Physics ($\mu=1-2\sqrt{\alpha}$) & 95.02 $\pm$ 0.19\% & \textbf{47.0 $\pm$ 6.4} \\
\bottomrule
\end{tabular}
\end{table}

\textbf{Speedup.} Physics reaches 90\% before baseline on 5/5 seeds; per-seed speedups are $2.08\times$, $2.86\times$, $2.58\times$, $2.46\times$, $1.71\times$ (mean $2.34\times$, min $1.71\times$). Friedman test across the three methods on epochs-to-90\%: $\chi^2 = 7.60$, $p = 0.022$; paired $t$-test baseline vs.\ physics: $t = 9.24$, one-sided $p = 4\times10^{-4}$ (two-sided $8\times10^{-4}$) (Wilcoxon $p = 0.031$, the exact floor at $n=5$).

\textbf{The accuracy cost is real.} Physics finishes below baseline on 5/5 seeds ($\Delta = 0.46$~pp, paired $t$-test $p = 0.009$). An earlier version of this paper described this gap as ``within seed-to-seed variance''; the multi-seed data refutes that wording. The deficit is a property of the schedule---near-critical damping suppresses the late-training oscillation that helps escape shallow minima---and it motivates the hybrid recipe of Section~\ref{sec:hybrid}, which removes it.

\textbf{Momentum alternatives do not explain the speedup.} On 3 seeds, Nesterov momentum ($\mu=0.9$) matches baseline accuracy (95.45--95.62\%) with no early speedup (94--106 epochs to 90\%), and heavy-ball $\mu=0.95$ is worse on both axes (94.85--94.98\%, 139--145 epochs).

\subsection{Short-Schedule Control}
\label{sec:control}

The comparison above holds the 200-epoch cosine schedule fixed for all methods, inviting the objection that a \emph{shorter} baseline schedule would reach 90\% quickly too. We test this directly: constant $\mu = 0.9$ with the cosine schedule compressed to $T \in \{60, 100\}$ epochs, three seeds each.

\begin{table}[ht]
\centering
\caption{Short-schedule control ($n=3$ seeds each) versus physics under the 200-epoch schedule ($n=5$). Compressed baselines trade final accuracy for speed; physics Pareto-dominates the accuracy-matched baseline.}
\label{tab:control}
\begin{tabular}{@{}lcc@{}}
\toprule
\textbf{Configuration} & \textbf{Epochs to 90\%} & \textbf{Best Test Accuracy} \\
\midrule
Baseline, $T{=}60$ & 38.7 $\pm$ 0.6 & 94.55 $\pm$ 0.02\% \\
\textbf{Physics, $T{=}200$} & \textbf{47.0 $\pm$ 6.4} & \textbf{95.02 $\pm$ 0.19\%} \\
Baseline, $T{=}100$ & 58.3 $\pm$ 3.2 & 94.98 $\pm$ 0.07\% \\
Baseline, $T{=}200$ & 107.8 $\pm$ 10.1 & 95.49 $\pm$ 0.08\% \\
\bottomrule
\end{tabular}
\end{table}

Compressing the schedule is not free: the $T{=}60$ baseline reaches 90\% in 39 epochs but caps at 94.55\%---$0.47$~pp \emph{below} physics and $0.93$~pp below the full baseline (computed from unrounded means; the rounded table values give $0.94$). At matched final accuracy the ordering is unambiguous: the $T{=}100$ baseline achieves the same $\approx$95.0\% as physics but needs 58 epochs to reach 90\% versus physics' 47. \textbf{Physics Pareto-dominates the accuracy-matched compressed baseline}, and Hybrid-90 (Section~\ref{sec:hybrid}) dominates the full 200-epoch baseline the same way: equal final accuracy, roughly half the epochs to 90\%. The speedup is a property of the momentum schedule, not of schedule length.

\medskip
The remainder of this section reports the original single-seed (seed 42) experiment in detail, which the multi-seed run replicates.

\subsection{Setup}

\textbf{Architecture.} ResNet-18~\cite{he2016deep} (11.2M parameters), modified for CIFAR-10~\cite{krizhevsky2009learning}: $3\times3$ initial convolution with stride~1 and padding~1 (replacing the standard $7\times7$), max-pooling replaced with identity, final FC layer mapping $512\to10$.

\textbf{Training.} 200 epochs, batch size 128, weight decay $5\times10^{-4}$, seed 42, cosine annealing from $\alpha_{\max}=0.1$ to $\alpha_{\min}=10^{-4}$. Data augmentation: random crop (32, padding 4) and random horizontal flip.

\textbf{Hardware.} NVIDIA Tesla P100 (16~GB), PyTorch 2.4.1, CUDA 12.1.

\textbf{Three conditions.} All conditions use the same cosine LR schedule; only the momentum schedule differs:

\begin{table}[ht]
\centering
\caption{Experimental conditions. All share the same cosine learning rate schedule.}
\label{tab:conditions}
\begin{tabular}{@{}lcc@{}}
\toprule
\textbf{Condition} & \textbf{Momentum Schedule} & \textbf{Free Params} \\
\midrule
Baseline & $\mu = 0.9$ (constant) & 1 \\
1cycle~\cite{smith2018superconvergence} & $\mu$: linear $0.95{\to}0.85$ over the first half, back to $0.95$ over the second & 2 \\
Physics (ours) & $\mu(t) = \text{clamp}(1 - 2\sqrt{\alpha(t)},\; 0.5,\; 0.99)$ & \textbf{0} \\
\bottomrule
\end{tabular}
\end{table}

\textbf{Clamping.} The physics momentum is clamped to $[0.5, 0.99]$ for numerical stability. At $\alpha_{\max}=0.1$, the raw formula gives $\mu \approx 0.37$, clamped to $0.50$; at $\alpha_{\min}=10^{-4}$, $\mu = 0.98$, within bounds. The clamp bounds are fixed a priori and were never tuned; we count them as part of the method definition rather than free parameters, and Table~\ref{tab:conditions} reflects tuned parameters. Incidentally, the exact discrete critical momentum for a unit-curvature quadratic is $(1-\sqrt{\alpha})^2 = 1 - 2\sqrt{\alpha} + \alpha$, which at $\alpha_{\max}=0.1$ equals $0.4675$ --- so the clamp at $0.50$ keeps the applied momentum closer to the exact discrete value than the first-order formula itself during the high-LR phase. This means the physics model is \emph{not} perfectly critically damped during the high-LR phase (approximately epochs 1--84, where $\alpha > 0.063$ under the 200-epoch cosine schedule): it is slightly underdamped due to clamping. However, the deviation is much smaller than the baseline's ($\Delta\mu = +0.13$ vs.\ $+0.53$ at epoch~1). See Figure~\ref{fig:momentum}.

\subsection{Results: Final Accuracy}

\begin{table}[ht]
\centering
\caption{Final accuracy (seed 42; all single-seed illustrations in this section use the committed raw data of the multi-seed run). The multi-seed analysis (Table~\ref{tab:multiseed}) shows the physics deficit is systematic ($-0.46$~pp, $p=0.009$), not seed noise.}
\label{tab:accuracy}
\begin{tabular}{@{}lcc@{}}
\toprule
\textbf{Method} & \textbf{Best Test Accuracy} & \textbf{Training Time (s)} \\
\midrule
Baseline ($\mu=0.9$) & \textbf{95.45\%} & 5{,}105 \\
1cycle (Smith) & 95.34\% & 5{,}095 \\
Physics ($\mu=1-2\sqrt{\alpha}$) & 95.04\% & 5{,}121 \\
\bottomrule
\end{tabular}
\end{table}

All three methods achieve comparable accuracy within 0.41 percentage points on this seed. Training time is identical---the physics schedule has no computational overhead.

\begin{figure}[t]
\centering
\includegraphics[width=\textwidth]{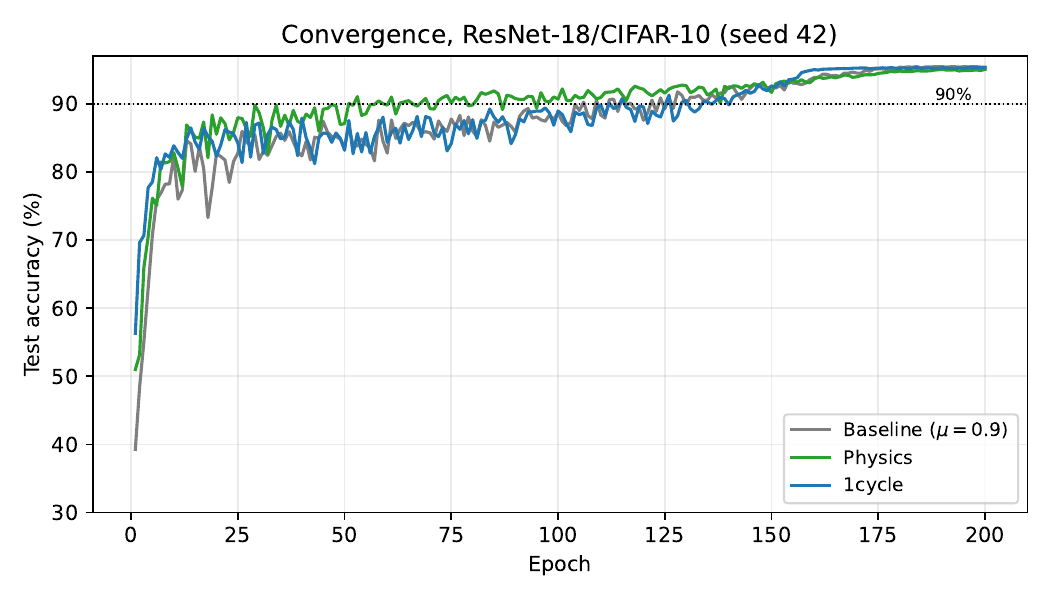}
\caption{Test accuracy during training (seed 42; multi-seed statistics in Table~\ref{tab:multiseed}). The physics method (green) converges fastest to intermediate thresholds, reaching 90\% at epoch~51 versus epoch~106 (baseline) and epoch~112 (1cycle). All methods converge to comparable final accuracy.}
\label{fig:convergence}
\end{figure}

\subsection{Results: Convergence Speed}

\begin{table}[ht]
\centering
\caption{Epochs to reach accuracy thresholds (seed 42). The physics method shows a $2.1\times$ speedup at 90\%; the multi-seed mean is $2.34\times$ (Table~\ref{tab:multiseed}).}
\label{tab:convergence}
\begin{tabular}{@{}lccccc@{}}
\toprule
\textbf{Method} & \textbf{80\%} & \textbf{85\%} & \textbf{90\%} & \textbf{92\%} & \textbf{95\%} \\
\midrule
Baseline & 10 & 26 & 106 & 140 & 174 \\
1cycle & 6 & 14 & 112 & 147 & 160 \\
\textbf{Physics} & \textbf{7} & \textbf{13} & \textbf{51} & \textbf{101} & 190 \\
\bottomrule
\end{tabular}
\end{table}

The physics method reaches 90\% accuracy at epoch~51---a \textbf{$2.1\times$ speedup} over baseline (epoch~106) and $2.2\times$ over 1cycle (epoch~112). This is consistent with the oscillator model's prediction that critical damping produces the shortest settling time. Note the plateau cost is visible on this seed too: physics reaches 95\% last (epoch~190).

\begin{table}[ht]
\centering
\caption{Early training accuracy at selected epochs.}
\label{tab:early_acc}
\begin{tabular}{@{}lccc@{}}
\toprule
\textbf{Method} & \textbf{Epoch 1} & \textbf{Epoch 10} & \textbf{Epoch 50} \\
\midrule
Baseline & 39.27\% & 82.36\% & 83.62\% \\
1cycle & \textbf{56.31\%} & \textbf{83.83\%} & 83.17\% \\
Physics & 50.99\% & 82.82\% & \textbf{87.07\%} \\
\bottomrule
\end{tabular}
\end{table}

\subsection{Prediction Verification}

The three qualitative predictions hold as follows:
\begin{enumerate}[leftmargin=*,topsep=2pt,itemsep=1pt]
    \item \textbf{Fastest early convergence---confirmed.} Physics reaches 90\% in approximately half the epochs of baseline (multi-seed: $2.34\times$).
    \item \textbf{Reduced oscillation---partially confirmed.} Test accuracy is not strictly monotone for any method at this scale; physics shows smaller epoch-to-epoch oscillation than baseline during epochs 1--60 (mean $|\Delta\text{acc}|$ 2.23 vs.\ 2.57~pp; largest single-epoch drop 5.25 vs.\ 7.22~pp on seed 42).
    \item \textbf{Competitive final accuracy---confirmed with a caveat.} The single-seed gap looks like noise, but the multi-seed analysis shows a systematic $-0.46$~pp deficit ($p=0.009$), which the hybrid recipe removes (Section~\ref{sec:hybrid}).
\end{enumerate}

\subsection{Damping Regime Analysis}

\begin{figure}[t]
\centering
\includegraphics[width=\textwidth]{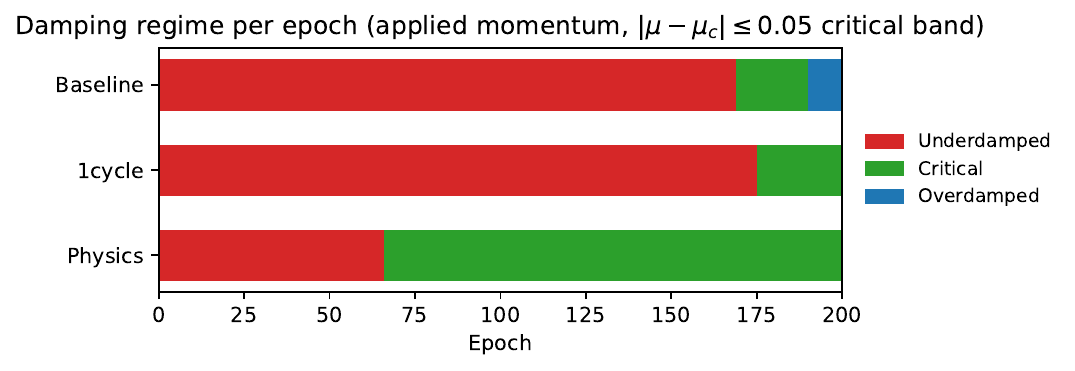}
\caption{Damping regime classification across 200 epochs. Red = underdamped ($\mu > \mu_c + 0.05$), green = critically damped ($|\mu - \mu_c| \leq 0.05$), blue = overdamped ($\mu < \mu_c - 0.05$). Classification uses the applied (clamped) momentum. The baseline is underdamped for 85\% of training; physics is mildly underdamped for the first 66 epochs of the clamped phase (the clamp is active through epoch 84) and critically damped thereafter, never overdamped.}
\label{fig:regimes}
\end{figure}

For each epoch, we classify the regime by comparing the actual momentum $\mu$ to the critical value $\mu_c = 1 - 2\sqrt{\alpha}$, using a tolerance of $|\mu - \mu_c| \leq 0.05$ for the critical band.

\begin{table}[ht]
\centering
\caption{Damping regime classification across 200 training epochs.}
\label{tab:regimes}
\begin{tabular}{@{}lccc@{}}
\toprule
\textbf{Method} & \textbf{Underdamped} & \textbf{Critical} & \textbf{Overdamped} \\
\midrule
Baseline ($\mu=0.9$) & 169 (84.5\%) & 21 (10.5\%) & 10 (5\%) \\
1cycle (Smith) & 175 (87.5\%) & 25 (12.5\%) & 0 (0\%) \\
\textbf{Physics} & \textbf{66 (33\%)} & \textbf{134 (67\%)} & \textbf{0 (0\%)} \\
\bottomrule
\end{tabular}
\end{table}

All rows classify the momentum \emph{actually applied} (i.e., after clamping for physics). The baseline operates in the underdamped regime for \textbf{85\% of training}---the standard configuration ($\mu=0.9$) spends the vast majority of training oscillating rather than converging optimally. Physics is never overdamped and is within the critical band for the final 134 epochs (67\%); during the clamped early phase it is mildly underdamped with $|\Delta\mu| \leq 0.13$---four times closer to critical than the baseline's $\Delta\mu = +0.53$ at epoch~1.

\textbf{Why this matters for diagnostics.} The damping regime analysis is not just a training diagnostic---it is the \emph{foundation} for the error-localization pipeline in Section~\ref{sec:exp2}. The fact that the baseline is heavily underdamped suggests that its learned representations may be suboptimal in specific layers, which we can now identify and correct.

\section{Generalization Across Architecture and Dataset}
\label{sec:generalization}

The multi-seed validation (Section~\ref{sec:exp1}) establishes the schedule on
ResNet-18/CIFAR-10. Two natural objections remain: the effect could depend on
the skip-connection structure of residual networks, and it could depend on the
dataset. We address both with a matched training protocol (200-epoch cosine
learning-rate schedule $0.1 \to 10^{-4}$, weight decay $5\times10^{-4}$, batch
size 128, three seeds); minor per-experiment implementation differences (cosine
phase convention, dataset normalization constants) are documented in the
released code.

\subsection{Architecture: VGG-16 Without Skip Connections}
\label{sec:gen-vgg}

Skip connections are a plausible confounder: they reshape the loss landscape
that the oscillator model damps, so a residual-only validation leaves open
whether the schedule exploits that structure. We train VGG-16(BN) adapted to
CIFAR-10 (global average pooling, single linear head).

\begin{table}[h]
\centering
\caption{VGG-16(BN)/CIFAR-10, three seeds. Epochs to 90\% test accuracy and
best accuracy; speedup is per-seed $\text{ep}_{\text{baseline}} /
\text{ep}_{\text{physics}}$.}
\label{tab:vgg}
\begin{tabular}{lccc}
\toprule
Method & Best acc (mean$\pm$std) & Epochs$\to$90\% & Speedup \\
\midrule
Baseline ($\mu=0.9$) & $94.01 \pm 0.28\%$ & $128.7 \pm 8.4$ & --- \\
Physics ($\mu = 1-2\sqrt{\alpha}$) & $93.64 \pm 0.07\%$ & $75.7 \pm 6.7$ & $1.72\times$ \\
\bottomrule
\end{tabular}
\end{table}

The speedup replicates on an architecture with no skip connections
($1.72\times$ to 90\% on average; per-seed $1.91/1.80/1.43\times$, 3/3 seeds,
paired $t=6.15$, $\mathrm{df}=2$, two-sided $p=0.026$), and the early-training
advantage is larger still (up to $4.9\times$ to 85\% on individual seeds). The
best-accuracy deficit of the pure physics schedule also replicates ($-0.07$ to
$-0.61$~pp), consistent with the ResNet-18 result: the fast-early/slow-top
profile is a property of the schedule, not of the architecture. We did not run
a hybrid arm on VGG; given that both the speedup and the deficit replicate, we
expect the Hybrid-90 recipe (Section~\ref{sec:hybrid}) to transfer, but this
remains to be validated directly.

\subsection{Dataset: CIFAR-100}
\label{sec:gen-c100}

On CIFAR-100, ResNet-18 does not reach 90\% top-1 accuracy, so we define the
milestone relative to the baseline: the first epoch at which a method reaches
95\% of the baseline's final best accuracy. Absolute threshold curves are
reported alongside in the released data.

\begin{table}[h]
\centering
\caption{ResNet-18/CIFAR-100, three seeds. Milestone = epochs to reach 95\% of
the baseline's final best top-1 accuracy.}
\label{tab:c100}
\begin{tabular}{lccc}
\toprule
Method & Best acc (mean$\pm$std) & Epochs$\to$milestone & Speedup \\
\midrule
Baseline ($\mu=0.9$) & $78.77 \pm 0.33\%$ & $152.7 \pm 1.2$ & --- \\
Physics ($\mu = 1-2\sqrt{\alpha}$) & $77.07 \pm 0.06\%$ & $134.3 \pm 3.5$ & $1.14\times$ \\
\bottomrule
\end{tabular}
\end{table}

On CIFAR-100 the effect is smaller: the early-training speedup persists
(reaching 60\% top-1 accuracy $2.2$--$2.9\times$ faster and 65\%
$2.3$--$3.9\times$ faster than the baseline), but the pure schedule pays a
larger final-accuracy cost ($-1.70$~pp on average, 3/3 seeds), which compresses
the accuracy-matched milestone gain to $1.14\times$ (0/3 seeds above
$1.3\times$). The milestone is threshold-sensitive in the direction one would
expect from the deficit: at 97\% of the baseline's best accuracy the pure
schedule is no faster ($0.99\times$ mean), and at $98\%$ it fails to reach the milestone on two of three seeds and is
slower than the baseline on the third --- at the top end, the accuracy cost
dominates the early speed. We therefore scope the speedup claim to the settings where it is
demonstrated --- CIFAR-10 across two architecture families --- and report the
CIFAR-100 result as a boundary of applicability. We note that this boundary
\emph{strengthens} the case for the hybrid recipe: the harder the dataset, the
more of the schedule's value lies in its early phase, which is exactly the
phase the hybrid keeps. Validating Hybrid-90 on CIFAR-100 is future work.

\section{Experiment 2: The Diagnostic Pipeline}
\label{sec:exp2}

This section describes the surgical-correction protocol and the original (version~2) experiment. \textbf{Reader note (v3):} the layer-\emph{selection} interpretation of this experiment is refuted by the exhaustive baseline of Section~\ref{sec:exhaustive}---the selected triple performs no better than a random 3-of-7 pick. The protocol itself, its reproducibility, and the attribution ranking remain valid and are prerequisites for Section~\ref{sec:exhaustive}.

\subsection{Hypotheses}

\textbf{H1: Error-specific gradient attribution localizes problem layers.} On misclassified test images, gradient norms should concentrate in specific layers rather than being uniformly distributed, producing a sparse layer-level diagnostic. \emph{Note:} gradient attribution is a well-established technique; what is new here is computing it \emph{exclusively on misclassified images} to identify ``sick'' layers rather than generally important ones.

\textbf{H2: Targeted correction is more efficient than full retraining.} Retraining only the identified problem layers should fix errors with $\geq$80\% compute savings.

\textbf{H3: Different damping trajectories produce different errors.} Models trained with different momentum schedules should make qualitatively different errors.

\subsection{Step 1 --- Scan: Cross-Model Error Analysis}

We compute per-image predictions for all three models on the 10{,}000-image CIFAR-10 test set:

\begin{table}[ht]
\centering
\caption{Cross-model error analysis on CIFAR-10 test set.}
\label{tab:errors}
\begin{tabular}{@{}lr@{}}
\toprule
\textbf{Category} & \textbf{Count} \\
\midrule
Common errors (all 3 methods wrong) & 215 \\
Only baseline wrong & 104 \\
Only physics wrong & 138 \\
Only 1cycle wrong & 96 \\
Physics correct, baseline wrong & 172 \\
\bottomrule
\end{tabular}
\end{table}

The 172 images correctly classified by the physics model but not by the baseline provide direct evidence that optimization trajectory affects learned representations (H3~confirmed). These are not random differences: they cluster around specific feature confusions, as shown below.

\subsection{Step 2 --- Localize: Gradient Attribution on Misclassified Images}

We compute per-layer gradient norms on the baseline model's 452 misclassified test images. Critically, we restrict the computation to \emph{errors only}---this asks ``which layers are most confused on the specific images the model gets wrong?'' rather than ``which layers are generally most important?''

\begin{table}[ht]
\centering
\caption{Per-layer gradient norms on misclassified images (baseline model). Layers above the median are flagged as problem layers.}
\label{tab:gradients}
\begin{tabular}{@{}llcl@{}}
\toprule
\textbf{Layer} & \textbf{Function} & \textbf{Grad Norm} & \textbf{Status} \\
\midrule
\texttt{layer3} & Mid-level features (shapes, parts) & \textbf{10.56} & Problem \\
\texttt{conv1} & Low-level features (textures, edges) & \textbf{9.14} & Problem \\
\texttt{layer2} & Early-mid features (patterns) & \textbf{8.58} & Problem \\
\texttt{layer1} & Very early features & 7.43 & Normal \\
\texttt{layer4} & High-level features & 7.38 & Normal \\
\texttt{bn1} & Batch normalization & 4.99 & Normal \\
\texttt{fc} & Final classifier & 0.86 & Normal \\
\bottomrule
\end{tabular}
\end{table}

\textbf{Result (H1 confirmed).} Gradient norms concentrate in three layers (\texttt{layer3}, \texttt{conv1}, \texttt{layer2}), with a $12.3\times$ ratio between the highest (\texttt{layer3}: 10.56) and lowest (\texttt{fc}: 0.86) layers, and a $1.15\times$ ratio between the third-highest problem layer and the fourth-highest normal layer. The distribution is sparse, not uniform---the pipeline produces a clear cut between ``sick'' and ``healthy'' layers.

\subsection{Step 3 --- Diagnose: Error Taxonomy}

The most frequent confusion pairs cluster around texture--shape confusions in the problem layers:

\begin{table}[ht]
\centering
\caption{Most frequent confusion pairs in the baseline model with layer attribution.}
\label{tab:confusions}
\begin{tabular}{@{}llll@{}}
\toprule
\textbf{True} & \textbf{Predicted} & \textbf{Attributed Layer} & \textbf{Root Cause} \\
\midrule
cat & dog & \texttt{layer3} & Face shape, furry texture \\
dog & cat & \texttt{layer3} & Reverse confusion \\
cat & bird & \texttt{layer2} & Compact silhouette \\
deer & dog & \texttt{conv1} & Brown texture $\to$ ``dog'' \\
horse & bird & \texttt{conv1} & Green background dominates \\
ship & airplane & \texttt{layer2} & Horizontal shape \\
frog & dog & \texttt{layer3} & Color confusion \\
\bottomrule
\end{tabular}
\end{table}

Each error has a \emph{causal explanation}: \texttt{conv1} confuses based on texture (brown $\to$ dog, green $\to$ bird); \texttt{layer2} confuses based on shape (compact silhouette, horizontal shape); \texttt{layer3} confuses based on mid-level features (face shape, fur, color). The diagnostic pipeline does not just identify \emph{that} errors exist---it explains \emph{why} each error occurs and \emph{which layer} is responsible.

\subsection{Step 4 --- Treat: Surgical Correction}

\textbf{Protocol.} We freeze all 120 parameter tensors in the baseline model, then unfreeze only the three problem layers (\texttt{layer3}, \texttt{conv1}, \texttt{layer2}; 35 tensors). We retrain for 30 epochs with physics momentum at $\alpha_{\max} = 0.01$ (10$\times$ lower than original training).

\begin{table}[ht]
\centering
\caption{Surgical correction results: retrain 3 of 7 layer groups for 30 epochs with physics momentum.}
\label{tab:correction}
\begin{tabular}{@{}lcc@{}}
\toprule
\textbf{Metric} & \textbf{Before} & \textbf{After} \\
\midrule
Accuracy & 95.48\% & 95.54\% \\
Total errors & 452 & 446 \\
Errors fixed & --- & 55 \\
New errors & --- & 49 \\
Net improvement & --- & +6 \\
\midrule
Compute (epochs $\times$ layers) & --- & 30 $\times$ 3 layers \\
vs.\ full retraining & --- & 200 $\times$ all layers \\
Compute savings (epoch-wise) & --- & 85\% \\
\bottomrule
\end{tabular}
\end{table}

\textbf{Result (H2 confirmed).} Surgical correction fixed 55 errors by retraining only 29\% of parameters (35/120 tensors) for 15\% of the original training duration---an 85\% epoch-wise compute saving versus full retraining, on 29\% of the parameters. The net improvement (+6) in this initial experiment is modest; a stronger variant is tested in Section~\ref{sec:exp3}.

\subsection{Step 5 --- Verify: Analysis of Side Effects}

The 49 new errors after correction are expected: retraining at the layer-group level affects many parameters simultaneously, and some previously correct features are perturbed. This can be mitigated through: (1)~per-residual-block correction instead of per-layer-group, (2)~elastic weight consolidation~\cite{kirkpatrick2017overcoming} to preserve correct features, and (3)~iterative diagnose--correct--verify cycles. Section~\ref{sec:exp3} shows that choosing layers more carefully (via iKFAD-guided selection) achieves the best net result (+22; 62 fixed, 40 new errors).

\section{Experiment 3: Cross-Optimizer Invariance and Diagnostic Comparison}
\label{sec:exp3}

This experiment tests the most important question for scaling: \emph{does the diagnostic generalize across optimizers?} If gradient attribution on misclassified images identifies the same problem layers regardless of how the model was trained, then the pipeline can be applied to any model---including LLMs trained with Adam variants.

\subsection{Setup}

We train five models under identical conditions (ResNet-18, CIFAR-10, 200 epochs, cosine LR, seed~42):
\begin{enumerate}[leftmargin=*,topsep=2pt,itemsep=1pt]
    \item \textbf{Baseline}: SGD + constant $\mu = 0.9$.
    \item \textbf{Physics (ours)}: SGD + $\mu(t) = 1 - 2\sqrt{\alpha(t)}$.
    \item \textbf{iKFAD}~\cite{karoni2026adaptive}: per-parameter adaptive friction $\xi_i$ tracking kinetic energy.
    \item \textbf{CD}~\cite{karoni2026adaptive}: cubically damped SGD ($c=0.1$).
    \item \textbf{Adam}~\cite{kingma2015adam}: $\text{lr}=10^{-3}$, cosine decay to $10^{-5}$, weight decay $5\times10^{-4}$.
\end{enumerate}

We compare two diagnostic methods and five correction strategies:
\begin{itemize}[leftmargin=*,topsep=2pt,itemsep=1pt]
    \item \textbf{Method~A (ours):} gradient attribution on misclassified images $\to$ layer-group surgical correction with physics momentum.
    \item \textbf{Method~B (Karoni et al.):} per-parameter friction $\xi_i$ from iKFAD training $\to$ rank layers by accumulated friction $\to$ layer-group surgical correction.
    \item \textbf{Methods~C, D:} per-parameter correction using top 30\% and 15\% highest-$\xi$ parameters.
    \item \textbf{Method~E:} cross-optimizer transfer---diagnose the Adam model with our gradient attribution, then correct with physics momentum.
\end{itemize}
All corrections retrain for 30 epochs with $\mu = 1 - 2\sqrt{\alpha}$.

\subsection{Results: Training Accuracy}

\begin{table}[ht]
\centering
\caption{Phase~1 accuracy: five training methods on ResNet-18/CIFAR-10 (200 epochs, seed~42).}
\label{tab:phase1_v6}
\begin{tabular}{@{}lcc@{}}
\toprule
\textbf{Method} & \textbf{Best Test Accuracy} & \textbf{Errors} \\
\midrule
CD---Cubic Damping (Karoni et al.) & \textbf{95.62\%} & 453 \\
Baseline ($\mu=0.9$) & 95.51\% & 453 \\
iKFAD (Karoni et al.) & 95.39\% & 468 \\
Physics ($\mu = 1 - 2\sqrt{\alpha}$) & 95.12\% & 496 \\
Adam ($\text{lr}=10^{-3}$) & 93.68\% & 647 \\
\bottomrule
\end{tabular}
\end{table}

All four SGD-momentum variants achieve comparable accuracy within a 0.50~pp range (95.12--95.62\%). Adam trails by ${\sim}1.8$~pp, consistent with the known generalization gap between SGD and Adam on image classification~\cite{wilson2017marginal}.

\subsection{Results: Cross-Optimizer Invariance (Key Finding)}

\begin{table}[ht]
\centering
\caption{Problem layers identified by each diagnostic method. The SGD--Adam overlap is 100\%.}
\label{tab:problem_layers_comparison}
\begin{tabular}{@{}lll@{}}
\toprule
\textbf{Method} & \textbf{Problem Layers} & \textbf{Basis} \\
\midrule
Ours (gradient attr.\ on SGD baseline) & \texttt{layer3, conv1, layer2} & Grad norm on errors \\
iKFAD (per-parameter friction $\xi$) & \texttt{bn1, layer3, layer2} & Accumulated friction \\
Ours (gradient attr.\ on Adam) & \texttt{conv1, layer2, layer3} & Grad norm on errors \\
\midrule
SGD--iKFAD overlap & \texttt{layer3, layer2} & 2/3 shared \\
\textbf{SGD--Adam overlap} & \textbf{\texttt{conv1, layer2, layer3}} & \textbf{3/3 shared (100\%)} \\
\bottomrule
\end{tabular}
\end{table}

\textbf{This is the central finding of the diagnostic track.} Gradient attribution on misclassified images identifies \emph{exactly the same three problem layers} on the SGD baseline and the Adam model---100\% overlap. This is not obvious \emph{a priori}: SGD and Adam traverse fundamentally different optimization trajectories, use different update rules, and produce models with different error counts (453 vs.\ 647). Yet the architectural bottlenecks that cause errors are the same.

The implication is that gradient attribution on misclassified images measures a property of the \emph{architecture}---specifically, which layers have insufficient capacity or receive insufficient gradient signal for the hardest examples---rather than an artifact of the optimizer. The diagnostic pipeline is therefore \textbf{optimizer-agnostic}, which is essential for applicability to large models trained with Adam variants.

Note the difference from iKFAD: our gradient attribution identifies \texttt{conv1} (first convolutional layer, learnable feature detectors), while iKFAD identifies \texttt{bn1} (batch normalization, only scale/shift parameters). Since \texttt{conv1} contains the features responsible for texture--shape confusions, the gradient-attribution diagnosis is more causally meaningful.

\subsection{Results: Surgical Correction Comparison}

\begin{table}[ht]
\centering
\caption{Surgical correction comparison: five strategies applied to the baseline model (Methods~A--D) and Adam model (Method~E). All corrections use physics momentum for 30 epochs.}
\label{tab:diagnostic_comparison}
\begin{tabular}{@{}lcccccc@{}}
\toprule
& & \textbf{A: Ours} & \textbf{B: iKFAD} & \textbf{C:} $\boldsymbol{\xi}$\textbf{-30\%} & \textbf{D:} $\boldsymbol{\xi}$\textbf{-15\%} & \textbf{E: Adam} \\
& \textbf{Before} & \textbf{(layers)} & \textbf{(layers)} & \textbf{(params)} & \textbf{(params)} & \textbf{(cross-opt)} \\
\midrule
Accuracy & 95.47\% & 95.54\% & \textbf{95.69\%} & 95.30\% & 95.37\% & 93.51\% \\
Errors fixed & --- & 62 & 62 & 42 & 46 & 146 \\
New errors & --- & 55 & \textbf{40} & 59 & 56 & 148 \\
Net improvement & --- & +7 & \textbf{+22} & $-$17 & $-$10 & $-$2 \\
\midrule
Diagnostic & --- & Grad attr. & Friction $\xi$ & Friction $\xi$ & Friction $\xi$ & Grad attr. \\
Granularity & --- & Layer & Layer & Param & Param & Layer \\
\bottomrule
\end{tabular}
\end{table}

Three findings emerge:

\paragraph{Layer-level correction outperforms parameter-level.} Methods~A and B (layer-level) both fix 62 errors, while Methods~C and D (parameter-level, top 30\% and 15\% by friction $\xi$) fix only 42 and 46 respectively. Moreover, parameter-level methods introduce \emph{more} new errors than they fix, yielding negative net improvement ($-$17 and $-$10). Simpler granularity wins: at the ResNet-18 scale, layer-group correction constrains the optimization sufficiently to avoid destabilizing adjacent features.

\paragraph{iKFAD layers produce fewer side effects.} Methods~A and B fix the same number of errors (62), but iKFAD-guided layer selection (Method~B) introduces only 40 new errors versus 55 for Method~A, achieving net +22 versus +7. The likely explanation: iKFAD identifies \texttt{bn1} instead of \texttt{conv1}, and correcting batch normalization parameters is lower-risk than modifying the first convolutional layer.

\paragraph{Adam cross-optimizer correction (Method~E).} The diagnostic transfers perfectly (100\% layer overlap). Correction fixes 146 errors---the largest number of any method---but also introduces 148 new errors (net $-$2). The Adam-trained weights occupy a different region of parameter space; the correction hyperparameters need optimizer-specific tuning. This is a \emph{proof of concept}: the diagnostic generalizes; the correction requires adaptation.

\section{Experiment 3b: The Exhaustive Layer-Selection Baseline}
\label{sec:exhaustive}

Experiments 2--3 reported that surgical correction of diagnostically selected layer groups improves the model: gradient attribution scored net $+6$ on the Experiment-2 checkpoint and $+7$ on the Experiment-3 checkpoint (two independently trained baselines with 452 and 453 test errors respectively), and the best variant --- iKFAD-guided selection of \texttt{bn1+layer2+layer3}, Method~B --- reported $+22$. In the exhaustive re-run below, that same combination under the standardized correction protocol scores $+9$, a first hint that individual deltas of this magnitude are dominated by run-to-run variation. Version~2 of this paper presented this as evidence that the diagnostic \emph{selects which layers to repair}. The natural objection---``would any three layers do?''---turns out to be answerable exactly: with 7 layer groups there are only $\binom{7}{3} = 35$ possible selections. We ran the identical correction protocol (30 epochs, physics momentum, same checkpoint) on \textbf{all of them}.

\begin{table}[ht]
\centering
\caption{Exhaustive layer-selection landscape: all 35 combinations of 3-of-7 layer groups, identical surgical protocol on the same baseline checkpoint (453 initial errors). Selected rows; full table in the appendix and released data.}
\label{tab:exhaustive}
\begin{tabular}{@{}clc@{}}
\toprule
\textbf{Rank} & \textbf{Combination} & \textbf{Net errors fixed} \\
\midrule
1 & conv1 + bn1 + layer3 & $+18$ \\
2 & bn1 + layer3 + layer4 & $+15$ \\
3 & conv1 + bn1 + layer2 & $+14$ \\
\textbf{11} & \textbf{conv1 + layer2 + layer3 (diagnostic pick)} & $\boldsymbol{+7}$ \\
33 & bn1 + layer1 + layer4 & $-13$ \\
34 & bn1 + layer4 + fc & $-14$ \\
35 & layer1 + layer4 + fc & $-20$ \\
\bottomrule
\end{tabular}
\end{table}

\textbf{The selection claim fails.} The diagnostic triple ranks 11th of 35 (exact $p = 11/35 = 0.31$)---not better than random selection. We note the absolute scale honestly: the entire 35-combination landscape spans roughly $\pm 0.2$~pp of test accuracy ($+18$ to $-20$ net errors out of 453 on the 10k test set); its value is methodological---it is the exact null distribution that layer-selection claims should be tested against---rather than practical. Replicates of the diagnostic arm vary by $\pm 1$ net error ($+6$ on the Experiment-2 checkpoint; $+7$ and $+8$ in two replicates on this section's checkpoint), consistent with the nondeterminism noted in Section~\ref{sec:multiseed}. The result is not measurement noise: an independent re-run of the diagnostic arm reproduced its net improvement ($+7$ vs $+8$), so the protocol is stable; the claim is what fails.

\textbf{What survives.} Two weaker properties of the gradient-norm ranking are real:
\begin{enumerate}[leftmargin=*,topsep=2pt,itemsep=1pt]
    \item \emph{A soft top-layer signal.} The 15 combinations containing \texttt{layer3}---the top-ranked layer---average $+6.2$ net errors fixed, versus $-1.4$ for the 20 combinations without it.
    \item \emph{A reliable harm-avoidance signal.} The bottom of the landscape is populated by combinations of the \emph{lowest}-ranked layers (\texttt{layer4}, \texttt{fc}, \texttt{bn1}-with-\texttt{layer4}): the anti-diagnostic pick loses $13$--$20$ net errors. The ranking correctly predicts what \emph{not} to touch.
\end{enumerate}

\textbf{Interpretation.} Gradient attribution on misclassified images measures where error-relevant computation concentrates, and its extremes are informative: the highest-norm layer is a good (not optimal) repair target, and low-norm layers are actively dangerous to retrain. But the fine ordering in the middle carries little information about repair outcomes, and combinations were never measured by the diagnostic at all---\texttt{bn1}, ranked 6th of 7, appears in all three top combinations. We conclude that layer-selection claims in surgical fine-tuning need combinatorial baselines, and we release the full landscape to make that comparison free for future work at this scale.

\section{Experiment 4: Hybrid Momentum Scheduling}
\label{sec:hybrid}
\label{sec:exp4}

Experiment~1 revealed a tension: physics momentum ($\mu = 1 - 2\sqrt{\alpha}$) converges to 90\% accuracy about twice as fast as baseline ($2.34\times$ multi-seed), but baseline ($\mu = 0.9$) reaches the final 95\% sooner (epoch~174 vs.\ 190 on seed~42). The physics schedule is near-critically damped, which suppresses the oscillation that helps escape shallow local minima in the late-training landscape. Baseline's underdamped regime acts as implicit exploration, beneficial when the remaining errors involve subtle feature distinctions.

This motivates a \textbf{hybrid schedule}: use physics momentum for fast early convergence, then switch to baseline momentum for the final push.

\subsection{Setup}

We compare five methods on the same setup (ResNet-18, CIFAR-10, 200 epochs, cosine LR, batch~128, seed~42):
\begin{enumerate}[leftmargin=*,topsep=2pt,itemsep=1pt]
    \item \textbf{Baseline}: constant $\mu = 0.9$.
    \item \textbf{Physics}: $\mu(t) = \text{clamp}(1 - 2\sqrt{\alpha(t)},\; 0.5,\; 0.99)$.
    \item \textbf{1cycle}~\cite{smith2018superconvergence}: $\mu$ linear $0.95{\to}0.85$ over the first half of training, back to $0.95$ over the second.
    \item \textbf{Hybrid-90}: physics momentum until test accuracy $\geq 90\%$, then $\mu = 0.9$.
    \item \textbf{Hybrid-ep52}: physics momentum until epoch~52 (fixed), then $\mu = 0.9$.
\end{enumerate}

The two hybrid variants test adaptive (accuracy-triggered) versus fixed (epoch-triggered) switching. The fixed switch point (epoch~52) was chosen, at design time, from the physics 90\% crossing observed in the run that preceded this experiment; on the committed Experiment-1 seed-42 data that crossing is epoch~51.

\subsection{Results}

\begin{table}[ht]
\centering
\caption{Hybrid momentum experiment: convergence milestones (epoch to reach threshold) and best accuracy. This run's log is 0-indexed, so milestones here read one lower than the 1-indexed convention used elsewhere in the paper.}
\label{tab:hybrid}
\begin{tabular}{@{}lccccccc@{}}
\toprule
\textbf{Method} & \textbf{Best Acc.} & \textbf{80\%} & \textbf{85\%} & \textbf{90\%} & \textbf{92\%} & \textbf{95\%} & \textbf{Switch} \\
\midrule
Baseline ($\mu=0.9$) & 95.33\% & 13 & 31 & 96 & 139 & 175 & --- \\
Physics ($\mu=1{-}2\sqrt{\alpha}$) & 95.11\% & \textbf{8} & \textbf{13} & \textbf{45} & \textbf{88} & 183 & --- \\
1cycle (Smith) & \textbf{95.36\%} & 17 & 43 & 85 & 137 & 179 & --- \\
Hybrid-90 & 95.30\% & 7 & 12 & 46 & 136 & \textbf{170} & ep47 \\
Hybrid-ep52 & \textbf{95.36\%} & 8 & 12 & 109 & 136 & 173 & ep52 \\
\bottomrule
\end{tabular}
\end{table}

\subsection{Analysis}

Three findings emerge:

\paragraph{Physics dominates early training.} Physics reaches 80\% at epoch~8 (vs.\ 13 baseline, 17 1cycle), 85\% at epoch~13 (vs.\ 31 baseline, 43 1cycle), and 90\% at epoch~45 (vs.\ 96 baseline, 85 1cycle). Both hybrid methods inherit this advantage.

\paragraph{Hybrid-90 reaches 95\% fastest.} After switching to baseline at epoch~47 (when physics first hit 90\%), Hybrid-90 reaches 95\% at epoch~170---5 epochs faster than baseline (175) and 13 epochs faster than pure physics (183). The adaptive switch combines the strengths of both regimes: near-critical damping for fast initial convergence, then underdamped exploration for the final refinement.

\paragraph{Hybrid-ep52 achieves highest tied accuracy.} Hybrid-ep52 matches 1cycle at 95.36\%---the best final accuracy---but reaches 90\% much later (epoch~109 vs.\ 46 for Hybrid-90). The fixed switch illustrates the fragility of epoch-triggered switching: this run peaked at 89.9\% at epoch~45 without holding 90\%, and the switch itself briefly cost ${\sim}8$~pp (80.5\% at epoch~52), so the first sustained 90\% crossing came only at epoch~109. The accuracy-triggered switch in Hybrid-90 avoids both failure modes.

\paragraph{Multi-seed validation.} Across five seeds, Hybrid-90 attains $95.45 \pm 0.05\%$ best accuracy (baseline: $95.49 \pm 0.08\%$; indistinguishable) while reaching 90\% in $45.2 \pm 9.5$ epochs ($2.4\times$ faster than the baseline's $107.8 \pm 10.1$), with switch epochs spanning 35--60. The five runs pool the two Hybrid-90 arms that ran alongside the Experiment-1 seed sweep (seeds 123 and 456; reported here rather than in Section~\ref{sec:multiseed}'s core comparison) with three reruns of the same recipe; the protocols match in all hyperparameters (see released code for minor implementation differences between experiment generations).

\paragraph{Interpretation.} The results confirm the oscillator-based intuition: physics momentum (near-critical damping) minimizes oscillation for fast descent in the high-curvature early landscape, while baseline momentum (underdamped) provides the perturbation needed to escape shallow basins in the flat late-training landscape. The hybrid schedule exploits both regimes in sequence, yielding the fastest path to 95\%.

\section{Implications for Large Models}
\label{sec:large_models}

\textbf{Reader note (v3):} this section originally argued for scaling the layer-\emph{selection} pipeline to LLMs. After the exhaustive baseline (Section~\ref{sec:exhaustive}), the honest version of that proposal is narrower: gradient attribution on failure cases may identify layers that are \emph{dangerous to modify} (harm avoidance) rather than optimal repair targets, and any LLM-scale selection claim needs a null far stronger than a few random picks. We keep the proposals below with that caveat.

\subsection{The Layer-Level Repair Problem}

Large language models exhibit failure modes---hallucination, bias, harmful outputs---that are currently addressed by retraining the entire model via RLHF or DPO~\cite{hierarchical2025}. This introduces an ``alignment tax'': fixing one failure mode may degrade performance elsewhere. Surgical fine-tuning~\cite{lee2023surgical} and hierarchical alignment~\cite{hierarchical2025} demonstrate that targeting specific layers can reduce interference.

The key missing ingredient is \emph{which layers to target}. Our diagnostic provides a principled answer: gradient attribution on failure cases identifies the layers most confused by the errors. The cross-optimizer invariance result (Section~\ref{sec:exp3}) suggests this diagnostic measures an \emph{architectural} property, which is essential for LLMs trained with Adam variants.

\subsection{Connection to Existing Methods}

Our gradient attribution identifies \emph{which layers} are responsible for specific errors. This is complementary to:
\begin{itemize}[leftmargin=*,topsep=2pt,itemsep=1pt]
    \item \textbf{ROME/MEMIT}~\cite{meng2022locating,meng2023memit}: localize \emph{factual associations} to MLP layers via causal tracing. Our method localizes \emph{error-causing computation} via gradient norms. The two could be combined: use gradient attribution to identify candidate layers, then apply ROME-style editing within them.
    \item \textbf{Representation Engineering}~\cite{zou2023representation}: identifies \emph{directions} in activation space encoding concepts (honesty, harmfulness). Our method identifies \emph{layers}. RepE's control vectors could be applied selectively to identified problem layers.
    \item \textbf{Surgical fine-tuning}~\cite{lee2023surgical}: demonstrates that partial retraining works. Our contribution is a diagnostic that tells you \emph{which part} to retrain.
\end{itemize}

\subsection{Proposed Experiments}

\textbf{Tier~1: LLM error localization.} Take a pre-trained LLM (e.g., Llama-3-8B) and a benchmark with known failure cases (TruthfulQA for hallucination, BBQ for bias). Compute per-layer gradient norms on failure cases. \emph{Hypothesis:} gradient norms will concentrate in a sparse set of transformer layers, analogous to our ResNet-18 finding.

\textbf{Tier~2: Surgical LoRA.} Apply LoRA adapters only to the layers identified in Tier~1 (``surgical LoRA'') versus full-model LoRA. \emph{Hypothesis:} surgical LoRA will fix a comparable number of errors with fewer side effects, at a fraction of the compute.

\textbf{Tier~3: Cross-architecture invariance.} Repeat Tier~1 on models with different architectures (Llama, Mistral, GPT-2-XL). \emph{Hypothesis:} error-causing layers will cluster at similar \emph{relative depths} (e.g., 30--50\% of model depth), suggesting a universal property of transformer architectures.

If confirmed at scale---which the present CIFAR-tier evidence cannot establish---these results would point toward gradient attribution on failure cases as a general-purpose diagnostic for large models.

\section{Discussion}

\subsection{A Diagnostic Framework, Not an Optimizer}

The oscillator model of SGD with momentum~\cite{qian1999momentum} has been known for 27 years. The formula $\mu = 1 - 2\sqrt{\alpha}$ is a direct consequence. Our contribution is not the formula but what it \emph{enables}: a validated schedule and hybrid recipe (Sections~\ref{sec:exp1} and~\ref{sec:generalization}), and a diagnostic pipeline whose real information content we quantify exactly. The exhaustive baseline (Section~\ref{sec:exhaustive}) shows the pipeline does \emph{not} select optimal layers to repair---its gradient-attribution ranking carries a soft positive signal at the top and a reliable harm-avoidance signal at the bottom. The cross-optimizer invariance (Table~\ref{tab:problem_layers_comparison}) still holds and elevates that (weaker) diagnostic from a training-specific tool to a property of the trained model itself.

\subsection{Hyperparameter Reduction}

A standard hyperparameter search treats $\alpha$ and $\mu$ as independent axes in a 2D grid. Eq.~\ref{eq:main} collapses this to a 1D curve: given any $\alpha$, the optimal $\mu$ is determined. This halves the search space for momentum-based optimizers---a useful byproduct of the physics, though not the primary contribution.

\subsection{Explanations for Model Errors}

Standard debugging identifies \emph{what} a model gets wrong. The diagnostic pipeline adds \emph{why}: the error taxonomy (Table~\ref{tab:confusions}) attributes each confusion pair to a specific layer and a specific feature confusion (texture, shape, color). This transforms debugging from a statistical exercise into an explanatory one.

\subsection{The Medical Analogy}

The five-step pipeline mirrors medical diagnostics:
\begin{enumerate}[leftmargin=*,topsep=2pt,itemsep=1pt]
    \item \textbf{Scan:} compute damping regime at each epoch---the ``vital signs.''
    \item \textbf{Localize:} gradient attribution on errors---the ``MRI.''
    \item \textbf{Diagnose:} map errors to feature confusions---the ``pathology report.''
    \item \textbf{Treat:} freeze healthy layers, retrain problem layers---the ``surgery.''
    \item \textbf{Verify:} before/after comparison on specific images---the ``follow-up.''
\end{enumerate}

\subsection{Layer-Level vs.\ Parameter-Level Correction}

Experiment~3 provides evidence that layer-group granularity outperforms per-parameter selection (Table~\ref{tab:diagnostic_comparison}). Both layer-level methods fix 62 errors; parameter-level methods fix only 42--46 with negative net improvement. We hypothesize that individual parameter selection introduces too many degrees of freedom, causing the correction to overfit to specific errors while destabilizing adjacent features. Layer-group correction constrains the optimization sufficiently to avoid this.

\subsection{Implications for Model Auditing}

Given the exhaustive-baseline result, the auditing implication of this work is narrower but better supported than in version~2: gradient attribution on failure cases reliably identifies which layers \emph{not} to touch (the bottom of the ranking predicts the most harmful interventions, down to $-20$ net errors), while positive selection is no better than chance. For safety-critical repair, that harm-avoidance signal is arguably the more valuable half: it bounds the blast radius of an intervention. The cross-optimizer invariance means this signal extends to models trained with any optimizer.

\section{Limitations and Future Work}

\textbf{Schedule-length confound (resolved).} The $2.34\times$ figure is measured under a fixed 200-epoch schedule; the short-schedule control (Section~\ref{sec:control}) shows the schedule-robust version of the claim: physics Pareto-dominates the accuracy-matched compressed baseline ($T{=}100$: 58 vs.\ 47 epochs to 90\% at equal $\approx$95.0\%), and beating physics on raw speed ($T{=}60$) costs $0.47$~pp of final accuracy.

\textbf{Sample sizes.} The core claims rest on $n=5$ seeds and the generalization/control arms on $n=3$; we therefore report per-seed values and ranges throughout and treat the quoted $p$-values as descriptive rather than confirmatory.

\textbf{Architecture and dataset coverage.} Our evidence covers ResNet-18 and VGG-16(BN) on CIFAR-10 and ResNet-18 on CIFAR-100 (Section~\ref{sec:generalization}), all with SGD+momentum under a cosine learning-rate schedule at $32\times32$ scale. On CIFAR-100 the pure schedule's final-accuracy cost grows and the accuracy-matched gain shrinks to $1.14\times$; validating the hybrid recipe there is future work. Transformers (ViT, language models) train predominantly with adaptive optimizers, where the momentum parameter is an EMA coefficient; extending critical damping to that setting is a distinct theoretical question and future work, as is the per-eigenmode schedule of Section~\ref{sec:spectral-grounding}.

\textbf{Hybrid recipe validated at $n=5$} (two arms run alongside the Experiment-1 seed sweep, pooled with three reruns)\textbf{.} Hybrid-90: best accuracy $95.45 \pm 0.05\%$ vs.\ baseline $95.49 \pm 0.08\%$ (indistinguishable), epochs to 90\%: $45.2 \pm 9.5$ vs.\ baseline $107.8 \pm 10.1$ ($2.4\times$). The switch epoch is stable across seeds (35--60).

\textbf{Surgical correction has thin margins everywhere.} Even the best combination in the exhaustive landscape ($+18$ of 453 errors) changes accuracy by only $0.18$~pp, and corrections introduce nearly as many errors as they fix (e.g., 49 new per 55 fixed for the diagnostic pick in Experiment~2). At this scale, surgical repair is a measurement instrument, not yet a practical technique.

\textbf{Global vs.\ per-layer momentum.} Equation~\ref{eq:main} applies a single $\mu$ to all layers. Each layer has its own effective curvature, suggesting a per-layer critical momentum. This connects to Karoni et al.~\cite{karoni2026adaptive}.

\textbf{Cross-optimizer correction tuning.} The diagnostic generalizes perfectly (100\% layer overlap SGD--Adam), but correction with physics momentum does not yet improve Adam-trained models (Method~E, net $-$2). Adapting correction hyperparameters to Adam's parameter scale is needed.

\textbf{Untested on large models.} Whether gradient attribution on failure cases produces sparse layer-level signal in transformer LLMs with billions of parameters is an open question (Section~\ref{sec:large_models}). \emph{What supports optimism:} the cross-optimizer invariance suggests the diagnostic measures architecture, not optimizer specifics, and ROME/MEMIT~\cite{meng2022locating} have demonstrated that factual knowledge in LLMs is layer-localized.

\textbf{Planned extensions:} (1)~the per-eigenmode schedule $\mu_k = 1 - 2\sqrt{\lambda_k \alpha}$ grounded in the spectral SDE of Olsen et al.~\cite{olsen2026spectra} (Section~\ref{sec:spectral-grounding}); (2)~validating Hybrid-90 on CIFAR-100, where the boundary identified in Section~\ref{sec:gen-c100} predicts it should recover most of the early-phase gain; (3)~extending critical damping to adaptive optimizers, where the momentum parameter is an EMA coefficient and the damping analysis differs; (4)~ImageNet-scale and transformer architectures.

\section{Conclusion}

The critical damping condition of the oscillator model of SGD with momentum~\cite{qian1999momentum} yields a momentum schedule with no tuned hyperparameters, and it works: $2.34\times$ faster convergence to 90\% accuracy under a fixed 200-epoch schedule across five seeds (one-sided $p = 4\times10^{-4}$), with a real and now-quantified final-accuracy cost ($-0.46$~pp, $p = 0.009$) that the Hybrid-90 recipe removes. These are the results of this paper that we expect to hold up, and they are immediately usable: given any learning-rate schedule, set $\mu(t) = 1 - 2\sqrt{\alpha(t)}$ until the model is near its accuracy plateau, then hold $\mu = 0.9$.

The paper's second lesson is methodological. Version~2 claimed that gradient attribution on misclassified images selects which layers to surgically repair; the exhaustive 35-combination baseline shows the selected triple is merely average (rank 11/35, $p = 0.31$), while the ranking's real information lives at its extremes---a soft signal at the top and a reliable harm-avoidance signal at the bottom. We were able to catch this only because the selection space at this scale is small enough to enumerate. Layer-selection claims---including in LLM-scale surgical fine-tuning and knowledge editing, where enumeration is impossible---deserve stronger nulls than a handful of random picks, because that is exactly the null that fooled us.

\subsection*{Reproducibility}

All code, raw per-epoch results, aggregation scripts, and the 35-combination landscape are publicly available at \url{https://github.com/anthroos/research} (directory \texttt{papers/paper2-beta-scheduling}). Experiments ran as Kaggle notebooks (mirrored in the repository together with their raw outputs). The schedule results (Sections~\ref{sec:multiseed}, \ref{sec:generalization}, \ref{sec:hybrid}) and the exhaustive baseline (Section~\ref{sec:exhaustive}) are fully recomputable from the committed raw data, including all figures (\texttt{experiments/figures/make\_figures.py}). The diagnostic-pipeline experiments of Sections~\ref{sec:exp2}--\ref{sec:exp3} predate this policy; their checkpoints and per-run outputs are reported as tabulated.

\subsection*{Acknowledgments}

Experiments were conducted on Kaggle Notebooks (NVIDIA Tesla P100, 16~GB). The author thanks the Kaggle platform for providing free GPU compute.

\bibliographystyle{plain}
\bibliography{references}

\appendix

\section{Epoch-by-Epoch Damping Scan (Baseline)}
\label{app:damping_scan}

\begin{table}[ht]
\centering
\caption{Detailed damping regime classification for the baseline model ($\mu=0.9$, cosine LR schedule with $\alpha_{\max}=0.1$).}
\label{tab:damping_scan}
\begin{tabular}{@{}ccccc@{}}
\toprule
\textbf{Epoch} & $\boldsymbol{\alpha(t)}$ & $\boldsymbol{\mu_{\text{actual}}}$ & $\boldsymbol{\mu_c}$ & \textbf{Regime ($\Delta\mu$)} \\
\midrule
1 & 0.10000 & 0.900 & 0.368 & Underdamped (+0.532) \\
20 & 0.09777 & 0.900 & 0.375 & Underdamped (+0.525) \\
50 & 0.08579 & 0.900 & 0.414 & Underdamped (+0.486) \\
100 & 0.05044 & 0.900 & 0.551 & Underdamped (+0.349) \\
150 & 0.01487 & 0.900 & 0.756 & Underdamped (+0.144) \\
170 & 0.00560 & 0.900 & 0.850 & Critical (+0.050) \\
180 & 0.00257 & 0.900 & 0.899 & Critical (+0.001) \\
200 & 0.00010 & 0.900 & 0.980 & Overdamped ($-$0.080) \\
\bottomrule
\end{tabular}
\end{table}

\section{Surgical Correction Training Log}
\label{app:correction_log}

\begin{table}[ht]
\centering
\caption{Training log for surgical correction of the baseline model (3 layers, 30 epochs, physics momentum).}
\label{tab:correction_log}
\begin{tabular}{@{}cccccc@{}}
\toprule
\textbf{Epoch} & $\boldsymbol{\alpha}$ & $\boldsymbol{\mu}$ & \textbf{Loss} & \textbf{Test Acc.} & $\boldsymbol{\Delta}$ \\
\midrule
1 & 0.01000 & 0.8000 & 0.0018 & 95.52\% & +0.04\% \\
10 & 0.00796 & 0.8216 & 0.0016 & 95.32\% & $-$0.16\% \\
20 & 0.00304 & 0.8898 & 0.0014 & 95.48\% & 0.00\% \\
30 & 0.00013 & 0.9775 & 0.0012 & 95.54\% & +0.06\% \\
\bottomrule
\end{tabular}
\end{table}

\section{Examples of Fixed Errors}
\label{app:fixed}

\begin{table}[ht]
\centering
\caption{Selected errors fixed by surgical correction, with layer attribution.}
\label{tab:fixed_examples}
\begin{tabular}{@{}clccl@{}}
\toprule
\textbf{\#} & \textbf{Description} & \textbf{Before} & \textbf{After} & \textbf{Attributed Layer} \\
\midrule
1 & Ship on bench & airplane & ship & \texttt{layer2} (horiz.\ shape) \\
2 & Horse in greenery & bird & horse & \texttt{conv1} (green texture) \\
3 & Cat on tree & bird & cat & \texttt{layer2} (silhouette) \\
4 & Dog close-up & cat & dog & \texttt{layer3} (face shape) \\
5 & Cat from above & dog & cat & \texttt{layer3} (furry texture) \\
6 & Frog & dog & frog & \texttt{layer3} (color) \\
7 & Deer & dog & deer & \texttt{conv1} (brown texture) \\
8 & Cat standing & dog & cat & \texttt{layer3} (cat/dog) \\
9 & Cat curled up & airplane & cat & \texttt{layer2} (gray shape) \\
10 & Cat lying down & bird & cat & \texttt{layer2} (compact shape) \\
\bottomrule
\end{tabular}
\end{table}

\end{document}